%% file: main.tex
\documentclass[10pt,twocolumn,letterpaper]{article}
\usepackage{cvpr}              

\input{preamble}

\definecolor{cvprblue}{rgb}{0.21,0.49,0.74}
\usepackage[pagebackref,breaklinks,colorlinks,allcolors=cvprblue]{hyperref}

\def\paperID{3887} 
\def\confName{CVPR}
\def\confYear{2025}

\title{Open-Vocabulary Functional 3D Scene Graphs for Real-World Indoor Spaces}

\author{
Chenyangguang Zhang$^{1,2}$ \qquad
Alexandros Delitzas$^{2,3}$ \qquad
Fangjinhua Wang$^{2}$ \qquad
Ruida Zhang$^{1}$ \vspace{3px}
 \\
Xiangyang Ji$^{1}$ \qquad
Marc Pollefeys$^{2,4}$ \qquad
Francis Engelmann$^{2,5}$
\vspace{4px}
\\
{ \small
$^{1}$Tsinghua University \quad
$^{2}$ETH Z\"urich \quad
$^{3}$MPI for Informatics \quad
$^{4}$Microsoft \quad
$^{5}$Stanford University \quad
}
}

\begin{document}
\maketitle

\input{sections/0_abstract}
\input{sections/1_introduction}
\input{sections/2_related_work}
\input{sections/3_problem_definition}
\input{sections/4_method}
\input{sections/5_experiments}

\input{sections/6_conclusion}
{
    \small
    \bibliographystyle{ieeenat_fullname}
    \bibliography{main}
}


\end{document}

%% file: preamble.tex
\usepackage[accsupp]{axessibility}  

\usepackage{caption}
\usepackage{multirow}
\usepackage{graphicx}
\usepackage{mdframed}
\usepackage{lineno}
\usepackage{arydshln}
\usepackage{overpic}

\newcommand{\name}{OpenFunGraph}
\newcommand{\datasetname}{FunGraph3D}

\newcommand{\graph}[0]{{\color{BrickRed}\pmb{\mathcal{G}}}}
\newcommand{\object}[0]{{\color{RoyalBlue}\pmb{\mathcal{O}}}}
\newcommand{\functional}[0]{{\color{orange}\pmb{\mathcal{R}}}}
\newcommand{\interactive}[0]{{\color{ForestGreen}\pmb{\mathcal{I}}}}

\newcommand\customparagraph[1]{\vspace{0.3em}\noindent\textbf{#1.}}

\newcommand{\boldparagraph}[1]{\vspace{0.5em}\noindent{\bf #1.}}

\renewcommand{\paragraph}[1]{\boldparagraph{#1}}

\usepackage{caption} 

%% file: sections/0_abstract.tex
\begin{abstract}
\vspace{-15px}
\\
We introduce the task of predicting \emph{functional} 3D scene graphs for real-world indoor environments from posed RGB-D images. Unlike traditional 3D scene graphs that focus on spatial relationships of objects, functional 3D scene graphs capture objects, interactive elements, and their functional relationships. Due to the lack of training data, we leverage foundation models, including visual language models (VLMs) and large language models (LLMs), to encode functional knowledge. We evaluate our approach on an extended SceneFun3D dataset and a newly collected dataset, \datasetname{}, both annotated with functional 3D scene graphs. Our method significantly outperforms adapted baselines, including Open3DSG and ConceptGraph, demonstrating its effectiveness in modeling complex scene functionalities. We also demonstrate downstream applications such as 3D question answering and robotic manipulation using functional 3D scene graphs.
See our project page at \href{https://openfungraph.github.io}{https://openfungraph.github.io}.
\end{abstract}

%% file: sections/1_introduction.tex
\vspace{-12px}
\section{Introduction}
\vspace{-3px}
\label{sec:introduction}

\begin{figure}
    \centering
    \begin{tabular}{cc}
         \footnotesize{Posed RGB-D Frames} & \footnotesize{Functional 3D Scene Graph} \\
         \hspace{0.2\columnwidth} & \hspace{0.6\columnwidth}  \\
    \end{tabular}
    \vspace{-15px}

    \includegraphics[width=\linewidth]{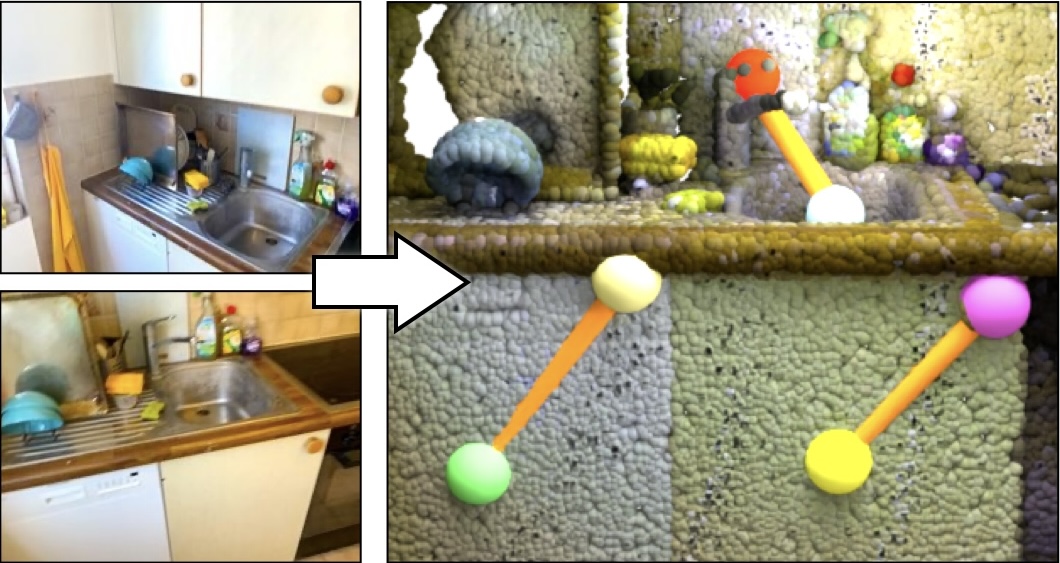}
    \includegraphics[width=\linewidth]{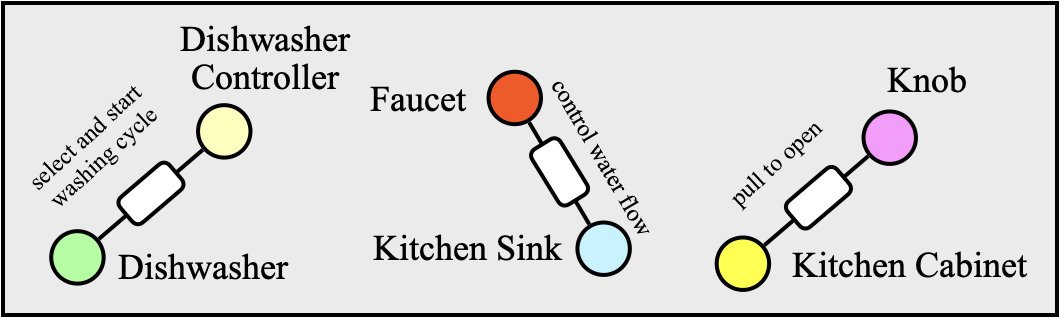}
    \caption{\textbf{Functional 3D Scene Graphs.}
Given an input sequence of posed RGB-D frames of an indoor environment, our method predicts a \emph{functional} 3D scene graph by detecting objects, identifying interactive elements, and inferring functional relationships. This enables the representation of interactions, functions, and scene dynamics, going beyond existing 3D scene graph methods that are constrained to \emph{spatial} relationships between static objects.
    }
    \label{fig:enter-label}
\vspace{-5px}
\end{figure}

This paper introduces \emph{functional} 3D scene graphs for real-world indoor spaces from posed RGB-D images. 3D scene graphs offer a lightweight, abstract representation for capturing the comprehensive semantic structure of an environment ~\cite{armeni20193d}. They support a variety of applications, including 3D scene alignment~\cite{sarkar2023sgaligner}, image localization~\cite{miao2024scenegraphloc}, graph-conditioned 3D scene generation~\cite{zhai2024commonscenes, dhamo2021graph}, as well as robotics navigation~\cite{werby2024hierarchical} and task planning~\cite{agia2022taskography, rana2023sayplan}.

Recent advances in 3D scene graph prediction
\cite{chen2024clip,koch2024open3dsg,gu2024conceptgraphs,armeni20193d,koch2024lang3dsg,wu2021scenegraphfusion,
rosinol20203d,rosinol2021kimera,wald2020learning}, have enabled exciting developments across multiple areas, including scene graph inference from 3D reconstructions \cite{wald2020learning, chen2024clip}, applications for robotic interactions \cite{gu2024conceptgraphs, wu2021scenegraphfusion}, online scene graph generation \cite{wu2021scenegraphfusion}, open-vocabulary 3D scene graphs \cite{koch2024lang3dsg, koch2024open3dsg} and large-scale, hierarchical scene graphs \cite{armeni20193d, rosinol20203d, rosinol2021kimera}.
The performance of recent scene graph methods also benefits from advancements in 3D scene understanding techniques~\cite{choy20194d, qi2017pointnet, schult2023mask3d}, which they rely on to extract objects and their semantics for modeling inter-object relationships.
However, existing 3D scene graph estimation methods~\cite{wald2020learning, koch2024lang3dsg, gu2024conceptgraphs, wu2021scenegraphfusion} face important limitations: graph nodes are typically restricted to \emph{objects}, and edges represent only \emph{spatial} relationships. For instance, edges primarily capture relative positions, such as \emph{‘the TV is mounted on the wall’} or \emph{‘the flower is placed on the table’}—information already implicitly encoded by object positions. Crucially, these methods lack representations of small interactive elements~\cite{delitzas2024scenefun3d} and their \emph{functional} relationships with other scene objects, which are essential for finer-grained interactions (\eg, flipping a switch to turn on a light), making them less suitable for higher-level \emph{functional reasoning}.
The key idea of this paper is to enhance 3D scene graphs with the capability to represent \emph{functional} relationships between objects and their interactive elements. A 3D scene graph that captures both functionalities and interactions opens up significant opportunities. For example, robotic agents can identify interactive elements and their functional relationships with objects to perform effective manipulation tasks, or graph-guided 3D scene generation methods~\cite{zhai2024commonscenes, dhamo2021graph} can, with this enriched representation, generate more dynamic and realistic environments by incorporating interactive elements and their effects.
However, creating functional 3D scene graphs is challenging.
Most importantly, there is a lack of training data to learn the complex functional relationships between objects and their interactive elements.
Unlike existing 3D scene graphs, functional 3D scene graphs require a more nuanced understanding of interactions and object affordances.
To address this, our approach implements an open-vocabulary pipeline for functional 3D scene graph inference, termed \emph{\name{}}, leveraging the extensive knowledge encoded within foundation models, including visual language models (VLM) and large language models (LLM). These models, pre-trained on vast amounts of multimodal data, include rich semantic information that can potentially be adapted for functional understanding. This leads us to the central question of this work:
\emph{``Can we harness foundation models to construct functional 3D scene graphs?"}

We evaluate our approach on two challenging datasets: an extended version of SceneFun3D~\cite{delitzas2024scenefun3d} with newly added functional relationship annotations, and \datasetname{}, a freshly collected real-world dataset featuring high-precision 3D laser scans, accurately registered 
To address these limitations, we introduce functional 3D scene graphs, which model objects, interactive elements, and their functional relationships within a unified structure (formally defined in Section \ref{section:FSG}). This representation extends traditional 3D scene graphs by incorporating interactive sub-parts alongside objects and representing functional relationships beyond simple spatial ones.
We argue that functional 3D scene graphs should possess the following characteristics. First, the representation should operate in an \emph{open-vocabulary} manner to enhance generalization and applicability. Second, it should be \emph{flexible}, allowing various attributes to be attached to nodes (\eg, sensor data, natural language captions, semantic features) and edges (\eg, relationship descriptions), thus ensuring adaptability for downstream applications.

In summary, our key contributions are:
\begin{itemize}
\item
We introduce functional 3D scene graphs that extend traditional 3D scene graphs by capturing functional relationships between objects and interactive elements.
\item
We propose a novel approach that leverages the knowledge embedded in foundation models, specifically VLMs and LLMs, to construct functional 3D scene graphs without task-specific training.
\item
We present a new real-world dataset, \datasetname{}, with ground-truth functional annotations, and demonstrate that our method outperforms adapted baselines, including Open3DSG and ConceptGraph.
\end{itemize}

\begin{figure*}[ht!]
    \centering
    \begin{tabular}{cccccc}
    &
         \footnotesize{\textbf{Node Detection} (Sec. \ref{section:candidates})} & &
         \footnotesize{\textbf{Node Description} (Sec. \ref{sec:node_description})} &
         \footnotesize{\textbf{Functional Edges} (Sec. \ref{sec:constructing_edges})}&\\
         \hspace{0.1\textwidth} &
         \hspace{0.17\textwidth} &
         \hspace{0.125\textwidth} &
         \hspace{0.1\textwidth} &
         \hspace{0.3\textwidth} \\
    \end{tabular}
    \vspace{-14px}

\begin{overpic}[abs,unit=1mm,width=\textwidth,]{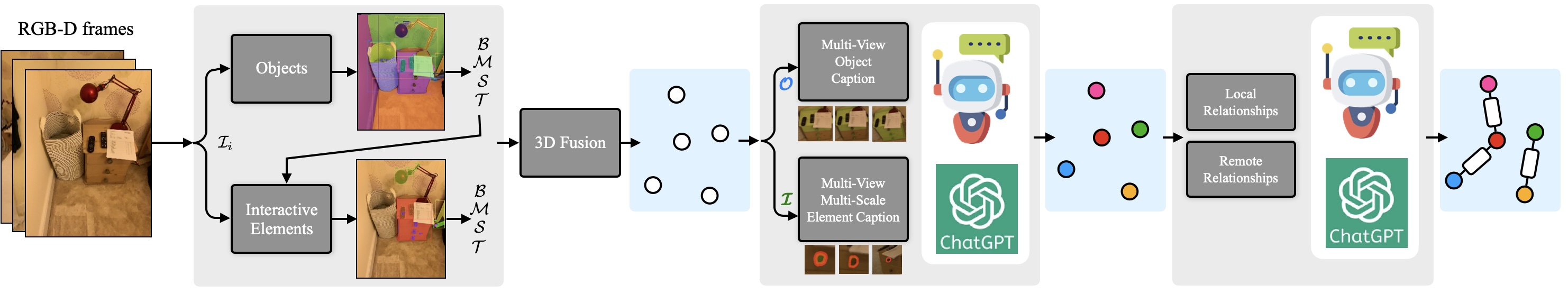} 
\put(73,26){\footnotesize $\object{} \cup \interactive{}$}
\put(119,26){\footnotesize $\object{} \cup \interactive{}$}
\put(161,26){\footnotesize $\graph{=}(\object{}, \interactive{}, \functional{})$}
\end{overpic}
    \caption{\textbf{Illustration of the \name{} architecture.}
    Given a sequence of posed RGB-D frames $\{(\mathcal{I}_i, \mathcal{D}_i)\}_{i=1}^{n}$, we use RAM++~\cite{zhang2024recognize} and GroundingDINO~\cite{liu2023grounding} to detect and segment objects $\object{}$ and interactive elemens $\interactive{}$, forming the node candidates of the functional 3D scene graph. Next, a mechanism using the large language model (LLM) GPT~\cite{achiam2023gpt} and the visual language model (VLM) LLAVA~\cite{liu2024llava} generates natural language descriptions $\mathcal{L}$ for each node. Finally, we infer functional relationships $\functional{}$ between objects $\object{}$ and interactive elements $\interactive{}$, represented as the edges in the functional 3D scene graph $\graph{}$.}
    \label{fig:pipeline}
\end{figure*}

%% file: sections/2_related_work.tex
\vspace{-7px}
\section{Related Work}
\vspace{-3px}
\label{section:rw}

\customparagraph{3D indoor scene understanding}
Many works concentrate on closed-set 3D semantic segmentation \cite{atzmon2018point,choy20194d,hu2021vmnet,hua2018pointwise,huang2023segment3d,landrieu2018large,weder2023alster,li2018pointcnn,qi2017pointnet,qi2017pointnet++,thomas2019kpconv,weder2024labelmaker} or instance segmentation \cite{engelmann20203d,han2020occuseg,hou20193d,takmaz20233d,jiang2020pointgroup,schult2023mask3d,vu2022softgroup,yue2023agile3d} on the existing 3D indoor scene understanding benchmarks \cite{dai2017scannet,chang2017matterport3d,ji2024arkit,sun2025nothing,rozenberszki2022language,armeni20163d,baruch2021arkitscenes,yeshwanth2023scannet++}.
With the development of foundation models, subsequent researches explore open-vocabulary 3D semantic segmentation~\cite{engelmann2024opennerf,kerr2023lerf,jatavallabhula2023conceptfusion,peng2023openscene,takmaz2023openmask3d,zhou2024feature,zuo2024fmgs,yilmaz2024opendas,qin2024langsplat,takmaz2025search3d}, and complex 3D visual language grounding tasks \cite{huang2022multi,yang2021sat,hsu2023ns3d,opencity3d2025,zhang2023multi3drefer,roh2022languagerefer,Parelli_2023_CVPR,Delitzas_2023_BMVC}.  
However, current studies mainly focus on object-level perception in indoor scene and seldom consider part-level interactive elements. 
Recently, SceneFun3D~\cite{delitzas2024scenefun3d} proposes a benchmark for functionality and affordance understanding, with exhaustive annotations of indoor interactive elements.
However, it does not provide the object annotations as well as the relationships between the elements and objects.
This work extends SceneFun3D by exploiting such relationships with functional 3D scene graphs.

\customparagraph{Affordance understanding}
Understanding affordance, \ie, properties of an environment to interact with, is a vital task in computer vision and robotics.
Existing learning-based methods usually take inputs such as images \cite{do2018affordancenet,zhai2022one}, videos \cite{fang2018demo2vec,nagarajan2019grounded,yoshida2024text} or 3D representations \cite{deng20213d,mo2022o2o,nagarajan2020learning,xu2022partafford}, and then predict affordance maps. 
Some works learn affordance from human-scene interaction demonstrations \cite{cho2024dense,banerjee2024introducing,ye2024g,fan2024hold,zhang2024moho,zhang2024ddf,ye2022s,chen2022alignsdf}. 
Nevertheless, existing works are often limited to object-level predictions and model affordances located on the corresponding objects. 
On the contrary, \name{} excavates all interactive elements at scene level, handling all kinds of functional relationships, especially those for remote operations. 

\customparagraph{3D scene graphs}
3D scene graph combines indoor entities into a unified structure and models inter-object relationships by building a graph of objects \cite{armeni20193d,rosinol20203d,rosinol2021kimera,wald2020learning,koch2024lang3dsg,wang2023vl,wu2021scenegraphfusion,wu2023incremental,zhang2021exploiting,zhang2021knowledge,takmaz2025search3d}. 
Functional 3D scene graph differs from the traditional 3D scene graph by adding interactive elements as nodes and modeling the functional relationships between objects and elements. 
Similarly, IFR-Explore~\cite{li2021ifr} tries to excavate inter-object functional relationships based on reinforcement learning in synthetic scenarios. 
However, it is hard to be applied in complex real-world scenes due to its closed-set setting, requirement of ground-truth instances, and lack of consideration on part-level elements. 
In this paper, we propose an open-vocabulary framework for functional scene graph inference in complex real-world scenes.
While there have been related efforts on open-vocabulary 3D scene graph generation, they are not well-suited for functional scene graph inference, particularly for interactive element recognition and functional relationship prediction. 
For example, Open3DSG~\cite{koch2024open3dsg} relies on object-level CLIP features~\cite{radford2021learning}.
It struggles with part-level interactive element recognition and is limited to inferring spatial relationships due to its design based on spatial-proximity edge feature distillation. ConceptGraph~\cite{gu2024conceptgraphs} uses a direct inference pipeline but focuses solely on object nodes and a narrow set of spatial relationships (\eg, \texttt{on}, \texttt{in}). 
In contrast, our approach introduces adaptive detection and description stages for both objects and interactive elements, alongside a sequential reasoning strategy for accurately modeling a wide range of functional relationships.

%% file: sections/3_problem_definition.tex
\vspace{-7px}
\section{Problem Formulation}
\vspace{-3px}
\label{section:FSG}

\paragraph{Functional 3D Scene Graphs}
We extend traditional 3D scene graphs \cite{gu2024conceptgraphs,koch2024open3dsg,wald2020learning} to facilitate their use in real-world scene interaction scenarios. Specifically, we introduce \emph{Functional 3D Scene Graphs}, a representation designed to enable functional reasoning by jointly modeling \emph{objects}, \emph{interactive elements} and their \emph{functional relationships}. We define a functional 3D scene graph as a directed graph $\graph{} = (\object{},\,\interactive{},\,\functional{})$ where $\object{}$ are the objects in the scene, $\interactive{}$ are the interactive elements and $\functional{}$ are the functional relationships which point from the interactive element nodes $\interactive{}$ to object nodes $\object{}$. Following the definition in \cite{delitzas2024scenefun3d}, we define interactive elements as components that agents interact with (\eg, handles, knobs and buttons) to trigger specific functions within the environment such as opening a cabinet or turning off a light. Additionally, functional relationships fall into two categories: \emph{local}, where the interactive element is part of the object (\eg, door-handle), or \emph{remote}, where the interactive element operates the object from a distance (\eg, TV-remote control).

\paragraph{Task definition} 
We formulate the following novel 3D scene understanding task: Given an input sequence of posed RGB-D frames $\{(\mathcal{I}_i, \mathcal{D}_i)\}_{i=1}^{n}$ 
of an unseen indoor environment, the task is to construct the functional 3D scene graph $\graph{}$ by inferring the functional relationships $\functional{}$ among the objects $\object{}$ and interactive elements $\interactive{}$ in the scene.

%% file: sections/4_method.tex
\vspace{-7px}
\section{Method}
\vspace{-3px}\label{section:method}

The goal of our method, \emph{OpenFunGraph}, is to predict the functional 3D scene graph of a 3D environment, by accurately detecting objects and interactive elements, and inferring the functional relationships among them in an open-vocabulary manner (Figure~\ref{fig:pipeline}). 
To overcome the challenge of limited training data, we harness the knowledge of foundation models~\cite{bommasani2021opportunities} to detect objects and interactive elements within the scene, describe them in natural language, and reason about their functional relationships.
In the detection stage (Section~\ref{section:candidates}), we follow a progressive strategy where we prompt the foundation model to systematically first identify objects and then transition to finer-grained interactive elements given the input image sequence. 
The 2D detection results are then fused across multiple viewpoints in 3D space, constructing an initial set of node candidates. 
Next, we utilize a VLM and an LLM to collaboratively generate multi-view aware natural language descriptions of the candidate nodes (Section~\ref{sec:node_description}). To construct the graph, we proceed with inferring the functional relationships, \ie, edges, among the object and interactive element nodes (Section~\ref{sec:constructing_edges}).
Specifically, we follow a sequential reasoning strategy, starting with local functional relationships (\eg, door - handle) and extending to remote functional relationships (\eg, TV -- remote control), by leveraging the common sense knowledge of VLMs and LLMs. This allows us to progressively build the scene's functional graph by incrementally establishing connections between nodes.

\subsection{Node Candidate Detection}
\label{section:candidates}

In the first stage, we detect objects and interactive elements in the scene to construct a set of node candidates.
We start by detecting 2D candidates on the input frames with a progressive foundation-model-based strategy that transitions from objects to finer-grained part-level interactive elements. Then, we associate and fuse the 2D detection results from multiple frames using geometric consistency, yielding the initial set of 3D node candidates.

\paragraph{Object candidates}
To identify object candidates $\mathcal{C}^{\mathcal{I}_i}_{o}$, we utilize RAM++~\cite{huang2023open,zhang2024recognize} to recognize objects in each input image $\mathcal{I}_i$, producing object tags $\mathcal{T}^{\mathcal{I}i}_{obj}$, such as `cabinet' or `door'. These object tags then serve as prompts for GroundingDINO~\cite{liu2023grounding}, which detects 2D bounding boxes $\mathcal{B}^{\mathcal{I}_i}$, segmentation masks $\mathcal{M}^{\mathcal{I}_i}$, and confidence scores $\mathcal{S}^{\mathcal{I}_i}$.

\paragraph{Interactive element candidates} 
Despite the increasing success of foundation models in detecting object instances within scenes, the development of prompting strategies for identifying smaller elements, including interactive object parts (\eg, knobs, handles), remains largely unexplored.
Here, we propose a simple yet effective strategy to generate suitable text prompts for GroundingDINO to improve the detection of small interactive parts. 
We ask the LLM GPT-4 to provide a list of potential interactive element tags corresponding to each object candidate tag $\mathcal{T}^{\mathcal{I}_i}_{obj}$.
We hold the valid object tags $\mathcal{T}^{\mathcal{I}_i}_{val}$ by filtering the cases where the LLM thinks the object is not interactable (\eg, wall, bed).
To create prompts for GroundingDINO, we concatenate $\mathcal{T}^{\mathcal{I}_i}_{val}$ (\eg, door) as assistive tags with the functional element tags (\eg, handle), forming prompts such as ``door. handle''.
Finally, we yield the interactive element candidates $\mathcal{C}^{\mathcal{I}_i}_{ie}$ in each input image $\mathcal{I}_i$ by maintaining the detections corresponding to the functional element tags.
Empirically, we observe that this approach leads to more accurate detection of small interactive parts. 
We support this observation with an ablation study in Section~\ref{section:ablation}.

\paragraph{3D candidate fusion}
After identifying the object and functional element candidates $\mathcal{C}_{obj}^{\mathcal{I}_i}$ and $\mathcal{C}_{ie}^{\mathcal{I}_i}$ in each image $\mathcal{I}_i$, we fuse their 2D segmentation masks using multi-view information to obtain the 3D node candidates of the graph. 
Following \cite{gu2024conceptgraphs}, we utilize the corresponding depth map $\mathcal{D}_i$ and camera projection matrix $\mathit{\Pi}_i$ to backproject the 2D mask to the 3D space, and merge them to receive the 3D object candidates $\mathcal{C}_{o}$ and interactive element candidates $\mathcal{C}_{ie}$. For each node candidate, we store the backprojected 3D point cloud $\mathcal{P}$ and 3D bounding box $\mathcal{B}$ along with the associated 2D image assets, \ie, images, masks, 2D bounding boxes and confidence scores.

\subsection{Node Candidate Description}
\label{sec:node_description}

We next outline the process of generating natural language descriptions $\mathcal{L}$ for each node by leveraging a combination of VLMs and LLMs. Precise language descriptions are critical for establishing functional relationships in the final phase.

\paragraph{Object candidates}
To generate natural language descriptions for each object candidate node, we first select the top $N_v$ views of each object, ranked by $\mathcal{S}^{\mathcal{I}_i} \times \frac{n_{\mathcal{P}^{\mathcal{I}_i}}}{n_{\mathcal{P}}}$, where $\mathcal{S}^{\mathcal{I}_i}$ is the 2D confidence score indicating the semantic confidence, while $n_{\mathcal{P}^{\mathcal{I}_i}}$ refers to the number of 3D points the view $\mathcal{I}_i$ contributes to the fused 3D pointcloud $\mathcal{P}$, presenting the geometric contribution of the view.
Each object is then cropped based on its bounding box $\mathcal{B}$, and a caption describing the object crop is obtained using LLAVA v1.6~\cite{liu2024llava,liu2024improved,liu2024visual}. Finally, to derive a unified language description for each object candidate, we employ GPT-4~\cite{achiam2023gpt} to summarize the multi-view LLAVA captions.

\paragraph{Interactive element candidates} 
Captioning small interactive elements poses additional challenges: the bounding box crops are considerably smaller, often containing only a few pixels, which hinders LLAVA's ability to generate accurate captions. To address this, we enlarge the bounding boxes by multiple scales to incorporate richer contextual visual information. 
Similar multi-scale approaches have been shown to be effective in \cite{kerr2023lerf,takmaz2023openmask3d}. To direct the VLM's attention to the interactive element within the expanded crop, we highlight the element with a red outline before passing it to LLAVA, as demonstrated in ~\cite{shtedritski2023does}. Finally, the multi-scale, multi-view captions are summarized into a single natural language description using GPT-4.

\subsection{Functional Relationships}
\label{sec:constructing_edges}

To model functional relationships between objects and interactive elements, we employ a sequential reasoning approach.
Drawing on the concept of \emph{Chain-of-Thought reasoning}~\cite{wei2022chain},
we decompose the task into a series of simpler steps rather than prompting the LLM to infer all possible element-object connections simultaneously.
Initially, we concentrate on identifying direct, local relationships between objects and elements that are rigidly connected (\eg, door -- handle). Once these relationships are established, we extend the search to remote relationships, where object-element pairs are functionally related but physically separated (\eg, TV -- remote control).

\paragraph{Local relationship reasoning}
First, we aim to construct the edges of the graph with local functional relationships, \eg, the keypanel of a microwave or the knob of a cabinet. 
A common characteristic of these cases is that objects and interactive elements are rigidly connected. 
To identify such cases efficiently, we first perform a spatial filtering process:
For each object node $\mathcal{C}_o^j$, we assess whether an element node $\mathcal{C}_{ie}^k$ has a significant spatial overlap.
Subsequently, we leverage the LLM's common sense knowledge to reason whether a local functional relationship between these two nodes is feasible.
To do this, we prompt the LLM with the language descriptions $\mathcal{L}^j$, $\mathcal{L}^k$ and 3D bounding boxes $\mathcal{B}^j$, $\mathcal{B}^k$ of $\mathcal{C}_o^j$ and $\mathcal{C}_{ie}^k$ respectively.
It is tasked with reasoning whether a local rigid connection between the interactive element (\eg, handle) and object (\eg, fridge) is feasible, and then generate a language description $\mathcal{L}^{k \to j}$ of the functional relationship (\eg, ``opens").
This step produces the subgraph of local connections $\hat{\graph{}^L}= \left(\object{}^L,\, \interactive{}^L,\, \functional{}^L\right)$.

\paragraph{Confidence-aware remote relationship reasoning}
In this step, we construct graph edges representing remote functional relationships, such as those between a ceiling light and its switch. Determining these remote relationships is challenging, as visual cues alone often do not fully clarify which interactive element controls which specific object. To address this, we introduce a confidence-aware reasoning strategy that assigns a confidence score to each inferred remote relationship. This approach enhances decision-making in real-world scenarios by enabling the agent to prioritize interactions with higher confidence scores.

First, we form an initial set of potential candidates for remote connections, by considering the interactive element nodes that remained unassigned from the previous stage.
To construct potential remote connections among the interactive elements and objects in the scene, we utilize the common sense knowledge of the LLM. 
Specifically, we provide the LLM with natural language descriptions $\mathcal{L}$ of the interactive element and object nodes, so that it can output a list of likely target objects that each interactive element could be functionally linked to.
Next, for each element-object pair, we employ the VLM to assess the feasibility of a functional connection. 
The visual input for this step is prepared by the top-1 views of the interactive element and object. 
The VLM can exploit useful information in the images of the element and object to generate descriptions for the feasibility assessment.
For example, it describes whether the appliance is physically plugged into the electric outlet, or whether the switch  is mount on the wall under the ceiling light.
The descriptions from all pairs are then provided to the LLM to form a global context, assisting it to assign a relative confidence score to each proposed connection and describe the nature of each relationship. 
This step outputs the subgraph of remote relations: $\hat{\graph{}^R}= \left(\object{}^R,\, \interactive{}^R,\, \functional{}^R\right)$.

\subsection{Final Graph Formation}

To construct the final graph, we combine the nodes and relationships identified in both the local and remote functional reasoning stages. The resulting predicted graph is formulated as
$\,
\hat{\graph{}} = \left(\object{}^L \cup \object{}^R,\, \interactive{}^L \cup \interactive{}^R,\, \functional{}^L \cup \functional{}^R\right)
$.

\begin{figure}
    \centering
    \includegraphics[width=\columnwidth]{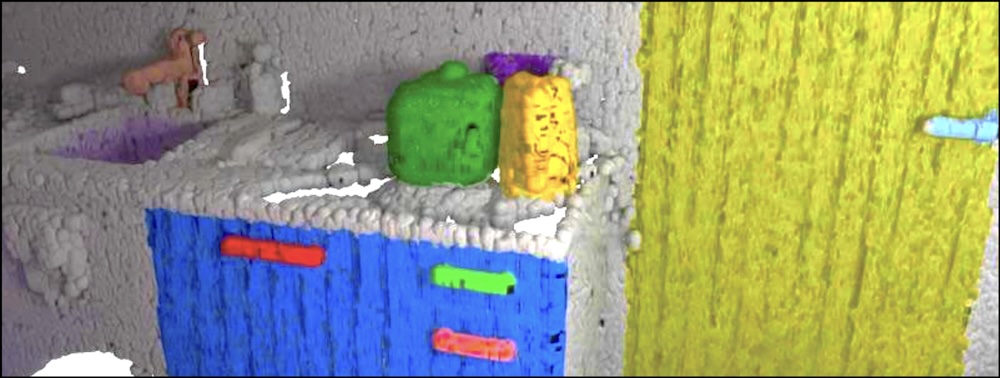}
    \includegraphics[width=\columnwidth]{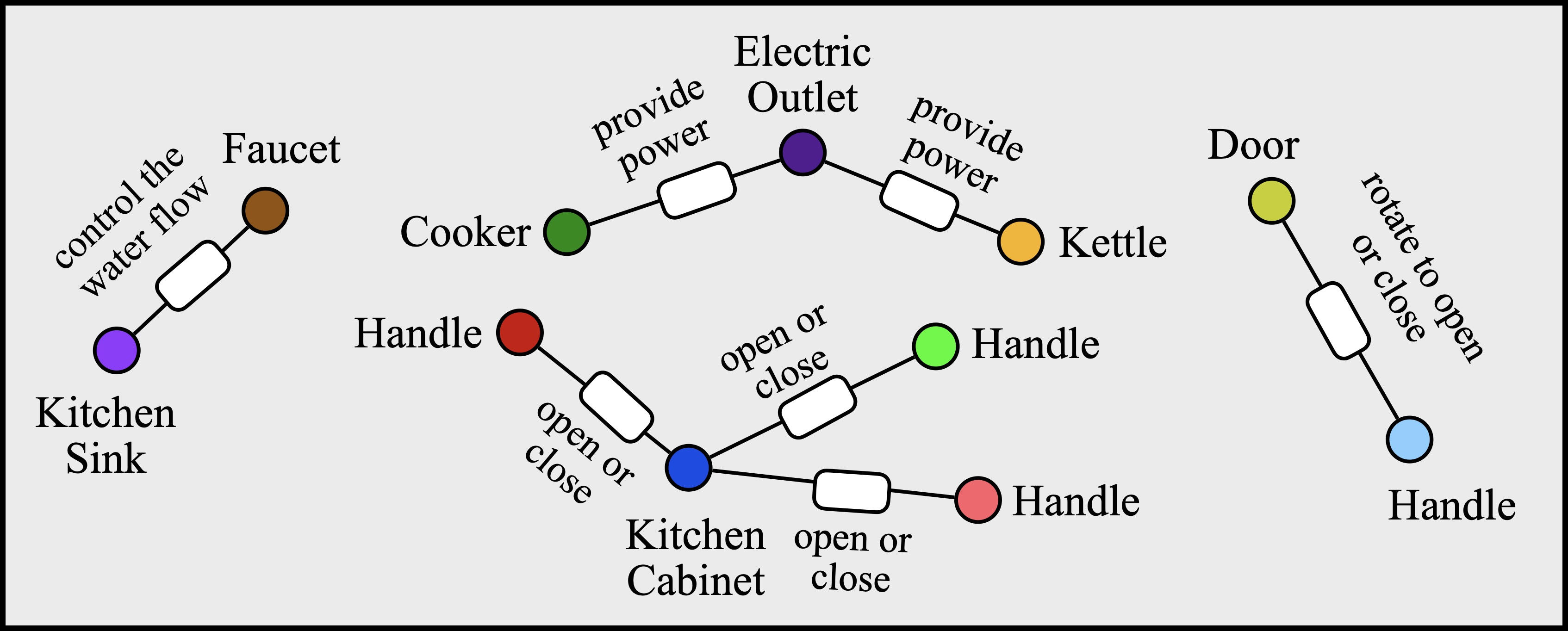}
    \includegraphics[width=\columnwidth]{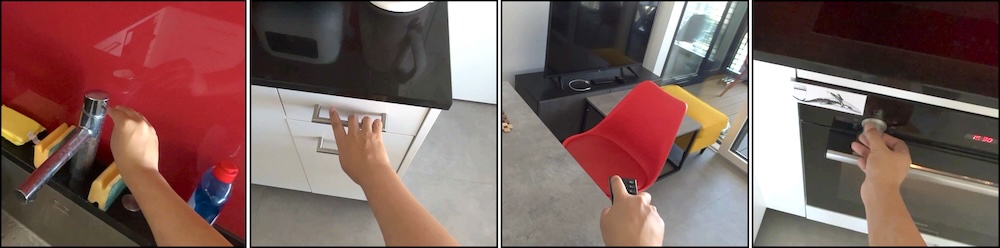}
    \caption{\textbf{Modalities of our \datasetname{} dataset.}
\emph{Top:} 3D scans from a Faro laser scanner, annotated with 3D object and interactive element masks.
\emph{Middle:} Ground truth functional 3D scene graphs.
\emph{Bottom:} Egocentric video capturing human-scene interactions.
}
    \label{fig:annotations}
\end{figure}

\begin{figure}
    \centering
    \includegraphics[width=\columnwidth]{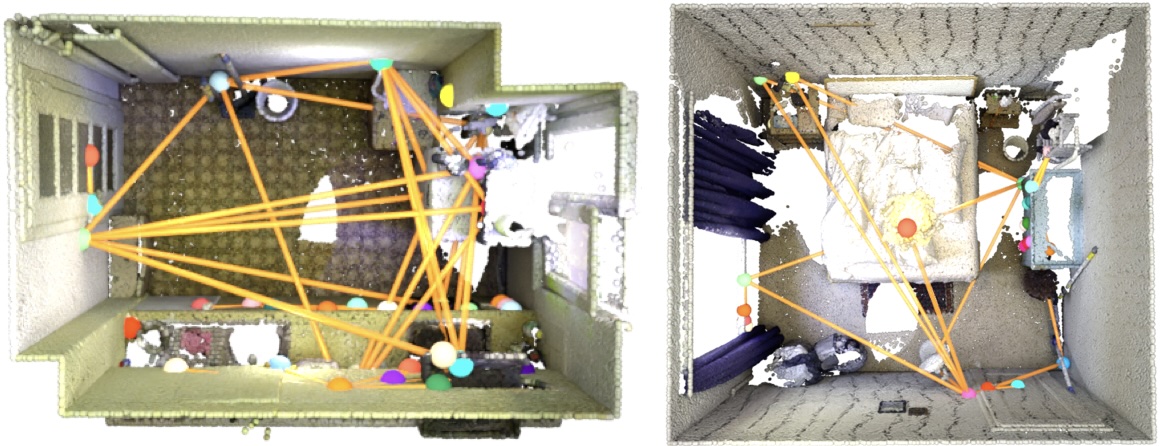}
    \caption{\textbf{Example scenes from our \datasetname{} dataset.} The dataset includes typical indoor environments such as living rooms, bedrooms, bathrooms, and kitchens.}
    \label{fig:scenes}
\end{figure}

\section{Data Collection}
\label{sec:data-collection}

Existing datasets of high-fidelity 3D indoor spaces focus primarily on understanding either 3D objects~\cite{baruch2021arkitscenes, yeshwanth2023scannet++} or 3D interactive elements~\cite{delitzas2024scenefun3d}. However, they lack ground-truth annotations of the functional relationships. In many cases, these relationships cannot be inferred from static visual observations alone but instead require video captures of physical interactions with the scene to determine which actions trigger specific responses. For example, a static 3D reconstruction cannot indicate which switch controls a particular light in a room with multiple switches and lights.
To systematically evaluate our method, we construct a novel dataset of 3D real-world indoor environments along with multi-sensor data (\ie, high-fidelity 3D reconstructions, consumer-device video captures, egocentric human-scene interaction videos) and functional 3D scene graph annotations. We outline the steps towards building this dataset, which we refer to as \emph{FunGraph3D} (Figure \ref{fig:scenes}). 

\paragraph{Laser scans} As illustrated in \cite{delitzas2024scenefun3d}, we highlight that laser scans can capture a higher level of 3D geometry details, such as small interactive elements (\ie, knobs, buttons), which is necessary for fine-grained scene understanding applications. To this end, we use a Leica RTC360 laser scanner to capture a high-resolution (5mm) 3D scan of the scene. To ensure high scene coverage during the capture, we place the scanner in multiple positions in the scene. We subsequently use the supporting software by Leica to fuse the multiple scans into a single one for the scene.

\paragraph{iPad video sequences} To enable scene understanding through multiple sensor data, we accompany the high-fidelity 3D reconstruction with RGB-D image information from a commodity device. Specifically, we capture multiple videos of the static scene with the camera of an iPad 15 Pro. 

\paragraph{Registration and alignment} 
To register the iPad video frames to the laser scan coordinate system, we build upon the COLMAP-based pipeline in \cite{yeshwanth2023scannet++}. 
Specifically, we run the COLMAP SfM pipeline~\cite{schoenberger2016sfm, schoenberger2016mvs} by augmenting the collection of real iPad frames with rendered pseudo images of the laser scan.
However, we notice that this pipeline leads to a large number of unregistered frames. To address this limitation, we incorporate the deep learning-based methods Superpoint~\cite{DeTone2017SuperPointSI} and Superglue~\cite{sarlin20superglue} for feature extraction and matching, leading to a more accurate registration result.
Afterwards, we utilize the optimized pose for each camera frame to render high-resolution depth maps for accurate back-projection from the iPad frames to the 3D space.

\paragraph{Egocentric videos} We include egocentric videos of property owners interacting with the environment using an Apple Vision Pro headset in our dataset. 
These videos facilitate accurate relationship labeling as they help clarify ambiguous connections among objects and interactive elements (\eg, which light switch controls the ceiling light). 

\paragraph{Annotation} For the annotation process, we extend the SceneFun3D annotation tool~\cite{delitzas2024scenefun3d} to construct the ground-truth functional 3D scene graphs. Annotators can navigate the 3D scene and annotate the instances of objects and interactive elements along with a free-form label. Annotators are also asked to connect the interactive element to the corresponding object that it controls and provide a description of their relationship. An example of the collected annotations is displayed in Figure~\ref{fig:annotations}. 

\paragraph{Statistics} \datasetname{} contains 14 in-the-wild scenes of various types (6 kitchens, 2 living rooms, 3 bedrooms and 3 bathrooms). In total, the dataset includes 201 interactive elements, 228 functional relationships and 146 objects of interest, along with open-vocabulary labels and relationships. 

%% file: sections/5_experiments.tex
\section{Experiments}
\input{figures/qualitative}

\input{tables/node_eval}

\subsection{Experimental Setup}\label{section:expsetup}
\vspace{-5px}
\customparagraph{Datasets}
To evaluate our method, we utilize the developed {FunGraph3D} dataset, described in Section~\ref{sec:data-collection}. 
Additionally, we use the {SceneFun3D} dataset~\cite{delitzas2024scenefun3d}, which provides high-resolution $5$\,mm laser scans of real-world environments along with iPad video sequences. 
Specifically, we randomly select 20 scenes (8 from the validation and 12 from the test split) and apply our annotation pipeline to annotate the functional 3D scene graph in each scene. 
Since we do not have physical access to the 3D environments, we restrict our evaluation to functional relationships that are visually unambiguous.
In total, 212 interactive elements, 195 functional relationships, and 105 corresponding objects are annotated for these scenes.

\customparagraph{Metrics}
To evaluate open-vocabulary functional 3D scene graphs effectively, a new quantitative metric is essential. Existing approaches, such as ConceptGraph~\cite{gu2024conceptgraphs}, rely on subjective human assessments, while Open3DSG~\cite{koch2024open3dsg} approaches evaluation as a label retrieval task, assuming all ground-truth nodes are known, an assumption that diverges from our real-world setting.
To address this, we extend the Open3DSG Recall@K metric~\cite{koch2024open3dsg} with a node detection component, using spatial overlap between predicted and ground-truth nodes, inspired by evaluation techniques on 2D scene graph generation~\cite{zhou2024openpsg,yang2022panoptic,lu2016visual,xu2017scene,yang2018graph}.
More specifically, our evaluation metric comprises two Recall@K scores: one for nodes, \ie, $\object$ and $\interactive$, and one for triplets, \ie, $(\object, \interactive, \functional)$.
For node evaluation, we preprocess all ground-truth labels to enable top-K retrieval, following Open3DSG~\cite{koch2024open3dsg}. 
A retrieval is considered successful if a ground-truth node has a non-zero 3D IoU with a predicted node and the ground-truth label ranks within the top-K retrievals based on cosine similarity of CLIP embeddings~\cite{radford2021learning} with the predicted label. 
We calculate overall node recall as $R_{no} = \frac{n^{re}_{no}}{n_{no}}$, where $n^{re}_{no}$ is the number of successfully retrieved ground-truth nodes, and $n_{no}$ is the total count of ground-truth nodes. 
Additionally, we assess recall for object and interactive element nodes separately, denoted as $R_{o} = \frac{n^{re}_o}{n_{o}}$ and $R_{ie} = \frac{n^{re}_{ie}}{n_{ie}}$, where \( n^{re}_o \) and \( n^{re}_{ie} \) are the counts of correctly retrieved objects and interactive elements and \( n_{o} \) and \( n_{ie} \) are their respective totals.
For triplet $(\object, \interactive, \functional)$ evaluation, we apply stricter criteria: a ground-truth triplet is successfully retrieved in the top-K only when all its components $\object$, $\interactive$ and $\functional$ are individually retrieved within the top-K. The retrieval process for
$\object$ and $\interactive$ follows the same approach as above.
To handle $\functional$, we preprocess all relationship annotations by generating BERT embeddings~\cite{devlin2018bert}, an approach effective for open-vocabulary predicates~\cite{koch2024open3dsg}.
Successful retrieval is based on cosine similarity between ground-truth and predicted BERT embeddings. 
Triplet recall is defined as $R_{tr}{=}\frac{n_{re}}{n_{tr}}$, where $n_{re}$ is the count of retrieved triplets, and $n_{tr}$ is the total count of ground-truth. 
We decompose triplet evaluation into node association ($R_{na}{=}\frac{n_{na}}{n_{tr}}$, with $n_{na}$ being the number of triplets retrieved only considering $\object, \interactive$), indicating node recognition, and edge prediction ($R_{ep}{=}\frac{n_{re}}{n_{na}}$), showing relationship inference given correct node associations.

\customparagraph{State-of-the-art comparisons}
We compare our approach against ConceptGraph~\cite{gu2024conceptgraphs} and Open3DSG~\cite{koch2024open3dsg}-based baselines. 
Two ConceptGraph-based baselines are reimplemented: ConceptGraph* modifies the original LLM prompts to infer functional relationships, rather than focusing on spatial relationships such as \texttt{in} or \texttt{on}. 
ConceptGraph* + IED further incorporates the proposed interactive element candidate detection (IED) from Section~\ref{section:candidates}, addressing ConceptGraph’s initial limitation in detecting small parts.
Both baselines use LLAVA v1.6 and GPT-4 for fair comparison with \name{}.
We also reimplement two Open3DSG-based baselines. 
Open3DSG* modifies the LLM prompts to output functional relationships instead of spatial relationships. 
Since Open3DSG baselines rely on ground-truth node instance segmentation for graph neural network inference, we implement Open3DSG*$^{\dagger}$, which uses \name{}'s fused 3D nodes for fair comparison. 
We report Recall@3 and Recall@10 for node metrics, and Recall@5 and Recall@10 for triplet metrics.

\subsection{Results}

Quantitative results are presented in Table~\ref{tab:node_eval} and \ref{tab:triplet_eval}. 
Overall, the \datasetname{} dataset poses a greater challenge than SceneFun3D~\cite{delitzas2024scenefun3d} due to its more complex scenes, which contain a higher number of objects and interactive elements.

\input{tables/triplets_eval}

\customparagraph{Node evaluation}
As shown in Table \ref{tab:node_eval}, \name{} surpasses ConceptGraph*~\cite{gu2024conceptgraphs} by 160\% on SceneFun3D and by 176\% in R@3 on \datasetname{}. 
ConceptGraph* primarily focuses on object perception, resulting in poor recall scores for interactive elements. 
With the added interactive element candidate detection (IED), ConceptGraph* + IED improves node recognition, but still falls short of \name{} by 22\% in R@3 on SceneFun3D, and 43\% in R@3 on \datasetname{}, thanks to the specified node description stage proposed in \name{}. 
Our approach also outperforms Open3DSG-based baselines, achieving 95\% and 29\% higher scores than Open3DSG*$^{\dagger}$ and Open3DSG* in R@3 on SceneFun3D, and 174\% and 66\% higher on \datasetname{}. 
The limited ability of Open3DSG-based methods to identify interactive elements arises from their focus on object-level features during training, whereas our approach employs a more practical open-vocabulary inference pipeline, free from these training constraints.

\customparagraph{Triplet evaluation}
Table \ref{tab:triplet_eval} shows triplet prediction results. 
On SceneFun3D and \datasetname{}, benefiting from accurate node recognition and the sequential reasoning strategy for functional inference, \name{} outperforms ConceptGraph* + IED by 76\% and 189\% in R@5, and Open3DSG*$^{\dagger}$ by 179\% and 308\%.
Notably, Open3DSG-based baselines struggle with functional relationships, as they rely on spatial edge features from adjacent instances. 
ConceptGraph-based methods, which prompt the LLM to predict all possible connections, also perform worse when compared to our sequential reasoning strategy due to the increased interpretive complexity imposed on the LLM.
Figure \ref{fig:vis} visualizes qualitative results for \name{}. 
In the left scene, our confidence-aware remote relationship reasoning successfully infers that the light switch is more likely to control the ceiling light rather than the two table light bulbs. 
In the right scene, the local functional relationship between the handle and the door is accurately identified. Additionally, the fan is most confidently inferred to be powered by the nearby electric outlet.

\subsection{Ablation studies}\label{section:ablation}

We ablate three key modules in our pipeline, \ie, the GroundingDINO prompts for interactive element candidate detection, sequential reasoning, and confidence-aware remote relationship reasoning, presented in Table~\ref{tab:ablation}.
The prompting strategy for GroundingDINO, which combines assistive object and element tags, proves effective. 
Using only element tags reduces node R@3 by 19\% and 10\%, as well as triplet R@5 by 20\% and 22\% on the two datasets respectively, due to incomplete detections. 
Replacing sequential reasoning with a direct approach, where the LLM infers functional relationships across all nodes, significantly reduces triplet reasoning performance (42\% and 32\% in triplet R@5 on SceneFun3D and \datasetname{} respectively). 
Sequential reasoning decomposes complex relationships into distinct types, making LLM processing easier.
Ablating confidence-aware remote relationship reasoning by randomly selecting connections, instead of using the highest-confident edge (\eg, choosing a random light for the switch instead of the most confident ceiling light), leads to a decrease in triplet R@5 by 7\% and 11\% on the two datasets respectively. 
This illustrates more reasonable edges are selected correctly in our mechanism by incorporating the common sense understanding of the foundation models.

\subsection{Downstream Applications}
\label{section:downstream}

\begin{figure}
    \centering
    \includegraphics[width=\linewidth]{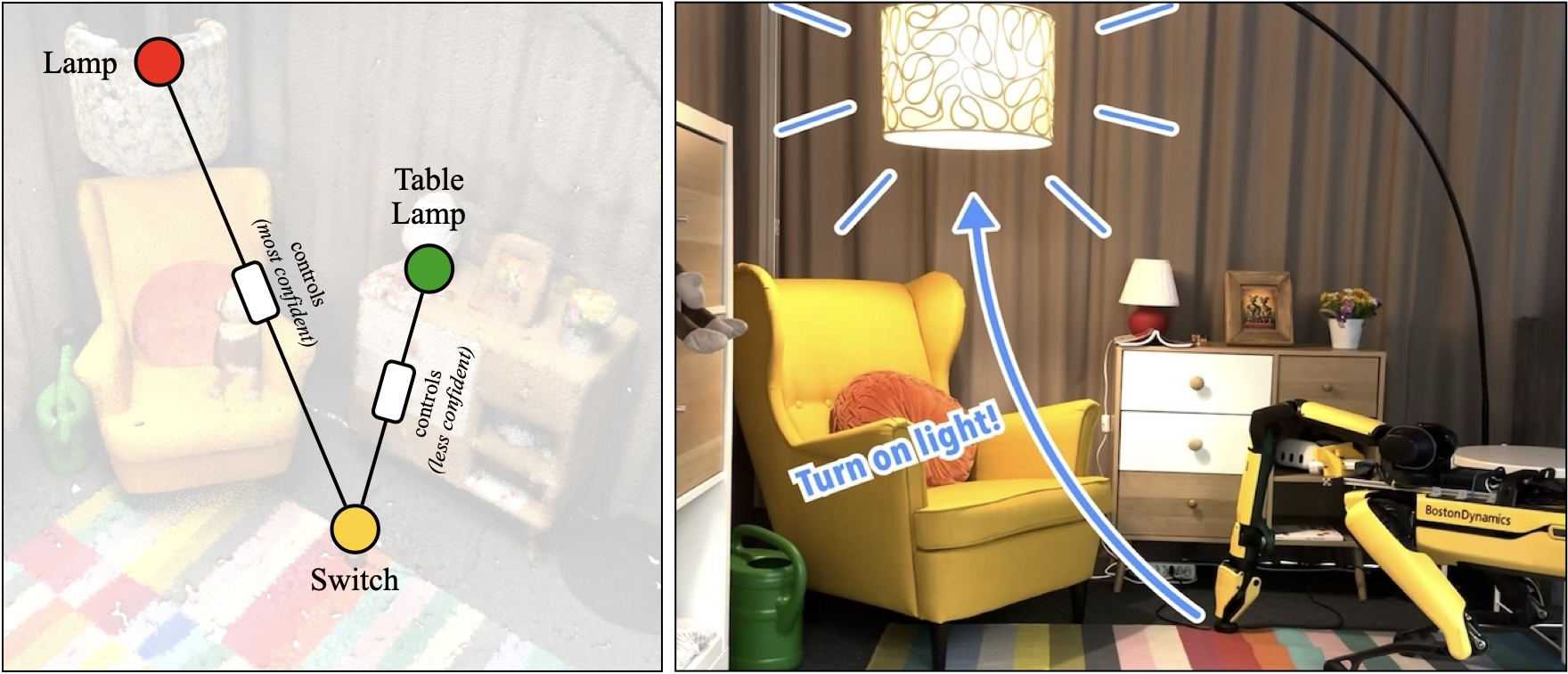}
    \caption{\textbf{Functional 3D Scene Graphs for Robotic Manipulation.}\\
    \emph{Left:} 3D scene and functional graph generated after querying `turning on the light.'
\emph{Right:} Robot interacting with scene elements as guided by the functional scene graph.}
    \label{fig:robot}
\end{figure}

We showcase the versatility of the proposed functional 3D scene graph representation in downstream applications that require complex reasoning about indoor functionalities and task-oriented interactions.

\paragraph{3D inventory question answering}
To enable functional reasoning, we convert the graph structure into a JSON list that the LLM can easily query. 
With this list, the LLM can answer questions such as “How can I turn on the ceiling light?”.
Using the functional 3D scene graph's nodes (objects, interactive elements) and edges (functional relationships), the LLM can provide responses such as “You can turn on the ceiling light using the light switch plate located at position [0.611, 0.113, 0.732]. From the provided JSON list, we can see the light switch plate with id 0 has the highest confidence level of 0.8 with the ceiling light fixture."

\paragraph{Robotic manipulation}
The functional 3D scene graph also supports robotic manipulation \cite{lemke2024spot,zurbrugg2024icgnet} for user queries that involve functional reasoning, as illustrated in Figure~\ref{fig:robot}. Similar to inventory question answering, the LLM queries the JSON list to locate the interactive element referenced in the query. 
The robot then navigates to and interacts with the element using the methods described in \cite{lemke2024spot}.

\input{tables/ablation}

%% file: figures/qualitative.tex
\begin{figure}
    \centering
\includegraphics[width=\columnwidth]{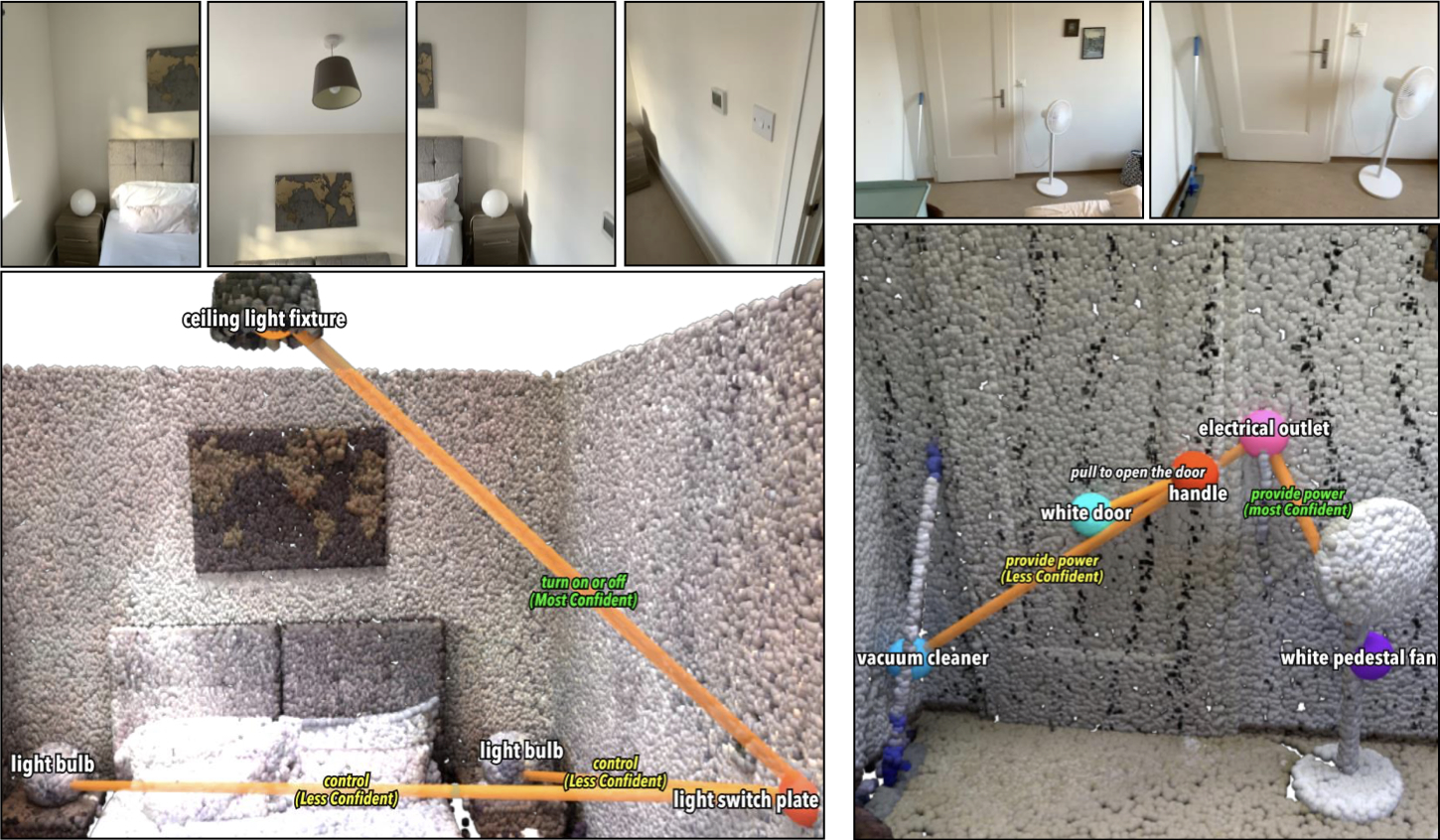}
    \caption{\textbf{Qualitative results.}
    \emph{Top:} input images.
    \emph{Bottom:} predicted functional 3D scene graph.
    Best seen zoomed in on a color screen.}
    \label{fig:vis}
\end{figure}

%% file: tables/node_eval.tex
\begin{table*}[t]
\setlength{\tabcolsep}{4px}
\centering
\resizebox{\linewidth}{!}{
\begin{tabular}{l cc cc cc c cc cc cc}
\toprule
& \multicolumn{6}{c}{\textbf{SceneFun3D}~\cite{delitzas2024scenefun3d}}
&& \multicolumn{6}{c}{\textbf{\datasetname{}}~(Ours)}
\\
\cmidrule{2-7}  \cmidrule{9-14}
& \multicolumn{2}{c}{Objects}
& \multicolumn{2}{c}{Inter. Elements}
& \multicolumn{2}{c}{Overall Nodes}
&
& \multicolumn{2}{c}{Objects}
& \multicolumn{2}{c}{Inter. Elements}
& \multicolumn{2}{c}{Overall Nodes}
\\
\textbf{Methods}
& R@$3$ & R@$10$ & R@$3$ & R@$10$  &R@$3$ & R@$10$
&
& R@$3$ & R@$10$ & R@$3$ & R@$10$  &R@$3$ & R@$10$
\\
\midrule
Open3DSG* \cite{koch2024open3dsg}
& $61.2$ & $70.7$ & $54.4$ & $61.8$ & $56.7$ & $64.7$ 
& & $50.9$ & $58.1$ & $21.8$ & $33.9$ & $33.4$ & $43.6$ \\
Open3DSG*$^{\dagger}$ \cite{koch2024open3dsg}
& $42.9$  & $50.0$ & $33.8$  & $38.3$  & $37.4$  & $43.0$
& &$30.9$  &$44.1$  &$13.0$  &$19.6$  &$20.2$ &$29.4$ \\
\addlinespace[1pt]\cdashline{1-14}\addlinespace[2pt]
ConceptGraph* \cite{gu2024conceptgraphs}
& $71.3$ & $77.1$ & $6.6$ & $8.6$ & $28.3$ & $31.4$
&
& $58.0$ & $66.3$ & $2.5$ & $4.1$ & $20.1$ & $25.2$\\
ConceptGraph*~\cite{gu2024conceptgraphs} + IED
& $71.3$ & $77.1$ & $53.1$ & $59.5$ & $60.1$  & $66.0$
&
 & $58.0$ & $66.3$ & $20.5$ & $33.4$ & $38.9$ & $45.0$ \\
\addlinespace[1pt]\cdashline{1-14}\addlinespace[2pt]
\name{} (Ours)
& $\mathbf{81.8}$ & $\mathbf{87.8}$ & $\mathbf{71.0}$ & $\mathbf{79.5}$ & $\mathbf{73.0}$ & $\mathbf{82.8}$ 
&
& $\mathbf{70.7}$ & $\mathbf{79.1}$ & $\mathbf{44.4}$ & $\mathbf{57.6}$ & $\mathbf{55.5}$  & $\mathbf{65.8}$ \\
\bottomrule
\end{tabular}
}
\caption{
\textbf{Node evaluation on the SceneFun3D~\cite{delitzas2024scenefun3d} and \datasetname{} datasets.}
* means to adapt the LLM prompts used for functional relationships inference.
IED refers to the interactive element candidate detection in Section \ref{section:candidates}.
$^{\dagger}$ refers to the usage of the \name{}'s fused 3D nodes rather than the ground-truth for fair comparison.
}
\label{tab:node_eval}
\end{table*}

%% file: tables/triplets_eval.tex
\begin{table*}[t]
\setlength{\tabcolsep}{4px}
\centering
\resizebox{\linewidth}{!}{
\begin{tabular}{l cc cc cc c cc cc cc}
\toprule
& \multicolumn{6}{c}{\textbf{SceneFun3D}~\cite{delitzas2024scenefun3d}}
&& \multicolumn{6}{c}{\textbf{\datasetname{}}~(Ours)}
\\
\cmidrule{2-7}  \cmidrule{9-14}
& \multicolumn{2}{c}{Node Assoc.}
& \multicolumn{2}{c}{Edge Pred.}
& \multicolumn{2}{c}{Overall Triplets}
&
& \multicolumn{2}{c}{Node Assoc.}
& \multicolumn{2}{c}{Edge Pred.}
& \multicolumn{2}{c}{Overall Triplets}
\\
\textbf{Methods}
& R@$5$ & R@$10$ & R@$5$ & R@$10$  &R@$5$ & R@$10$
&
& R@$5$ & R@$10$ & R@$5$ & R@$10$  &R@$5$ & R@$10$
\\
\midrule
Open3DSG* \cite{koch2024open3dsg}
& $47.2$ & $58.0$ & $69.2$ & $78.8$ & $32.7$ & $45.7$
& & $22.8$ & $36.7$ & $47.9$ & $55.9$ & $10.5$ & $20.0$ \\
Open3DSG*$^{\dagger}$ \cite{koch2024open3dsg}
& $ 33.6$ & $38.8$ & $64.4$ & $72.3$ & $21.6$ & $28.1$
& & $15.7$ & $24.2$ & $46.6$ & $55.7$ & $7.3$ & $13.5$ \\
\addlinespace[1pt]\cdashline{1-14}\addlinespace[2pt]
ConceptGraph* \cite{gu2024conceptgraphs}
& $5.6$  & $6.8$ &$80.2$  & $95.0$  & $4.7$  & $6.4$
&
& $1.9$ & $2.8$ &$51.5$ &$84.6$  & $1.1$ & $2.5$ \\
ConceptGraph*~\cite{gu2024conceptgraphs} + IED
&$45.4$  & $49.3$  & $75.6$ &$90.9$  & $34.3$  & $44.5$
&
 & $18.8$ & $22.8$  & $46.1$  & $79.7$  & $10.3$ & $18.9$ \\
\addlinespace[1pt]\cdashline{1-14}\addlinespace[2pt]
\name{} (Ours)
& $\mathbf{68.3}$ & $\mathbf{73.0}$ & $\mathbf{88.1}$ & $\mathbf{96.2}$ & $\mathbf{60.4}$ & $\mathbf{70.3}$ 
&
& $\mathbf{45.8}$ & $\mathbf{49.3}$ & $\mathbf{65.1}$ & $\mathbf{91.4}$ & $\mathbf{29.8}$  & $\mathbf{45.0}$ \\
\bottomrule
\end{tabular}
}
\caption{
\textbf{Triplet evaluation on the SceneFun3D~\cite{delitzas2024scenefun3d} and \datasetname{} datasets.}
All marks keep the same meaning with Table \ref{tab:node_eval}.
Node Assoc. refers to the node association metric while Edge Pred. means the edge prediction metric.}
\label{tab:triplet_eval}
\end{table*}

%% file: tables/ablation.tex
\begin{table}[t!]
\centering
\resizebox{\linewidth}{!}{
\begin{tabular}{lccccc}
\toprule
 & \multicolumn{2}{c}{\textbf{Overall Nodes}} && \multicolumn{2}{c}{\textbf{Overall Triplets}}\\
\cline{2-3} \cline{5-6}
 \textbf{Experiments} & R@3 & R@10 && R@5 & R@10 \\
\hline
w/o prompts for element detection &$59.3$  &$68.7$  &&$48.3$  &$59.9$ \\
w/o sequential edge reasoning*   & $\mathbf{73.0}$ & $\mathbf{82.8}$ && $34.8$ & $48.9$\\
w/o confidence-aware edge reasoning* & $\mathbf{73.0}$ & $\mathbf{82.8}$  && 56.0  & 65.1\\
Ours                        & $\mathbf{73.0}$ & $\mathbf{82.8}$ && $\mathbf{60.4}$ $\mathbf{70.3}$\\
\midrule
w/o prompts for element detection &$49.9$  &$59.1$  &&$23.1$  &$37.6$ \\
w/o sequential edge reasoning*   & $\mathbf{55.5}$ & $\mathbf{65.8}$ && $20.2$ & $33.8$\\
w/o confidence-aware edge reasoning* & $\mathbf{55.5}$ & $\mathbf{65.8}$ && 26.8 & 40.1\\
Ours                        & $\mathbf{55.5}$ & $\mathbf{65.8}$ && $\mathbf{29.8}$ & $\mathbf{45.0}$\\
\bottomrule
\end{tabular}
}
\caption{
\textbf{Ablation study} on SceneFun3D \cite{delitzas2024scenefun3d} (Top) and our \datasetname{} (Bottom).
Note that edge reasoning ($^*$) impacts only the triplet metric and does not affect node recognition performance.}
\label{tab:ablation}
\end{table}

%% file: sections/6_conclusion.tex
\section{Conclusion}
We introduce Functional 3D Scene Graphs, a novel representation that jointly models objects, interactive elements, and their functional relationships in 3D indoor environments. Our open-vocabulary pipeline leverages the common-sense knowledge of foundation models to infer functional 3D scene graphs and enable flexible querying. To support systematic benchmarking, we develop a high-fidelity dataset of real-world 3D indoor environments with multi-modal data and functional annotations. Experiments on this and existing datasets show that our method significantly outperforms baselines. We further demonstrate the versatility of our representation for downstream tasks such as 3D question answering and robotic manipulation.

\clearpage

\paragraph{Acknowledgments}
We would like to thank colleagues and friends who helped us capture the data of \datasetname{}: Christine Engelmann, Dominik Faerber, Elisabetta Fedele, Xudong Jiang, Xin Kong, Aoxue Liu and Houssam Naous. 
This work was supported by the Swiss National Science Foundation Advanced Grant 216260: ``Beyond Frozen Worlds: Capturing Functional 3D Digital Twins from the Real World''. AD is supported by the Max Planck ETH Center for Learning Systems (CLS) and FE by an SNSF PostDoc.Mobility Fellowship.